\documentclass[11pt]{article}

\usepackage[numbers]{natbib}

\usepackage{booktabs}       
\usepackage{amsfonts}       
\usepackage{amsmath}
\usepackage{graphicx}
\usepackage{amsthm}
\usepackage{amssymb}
\usepackage{multirow}
\usepackage{makecell}
\usepackage[vlined,linesnumbered,ruled,resetcount]{algorithm2e}
\usepackage[colorlinks,linkcolor=magenta,filecolor=blue,citecolor=blue,urlcolor=blue]{hyperref}
\usepackage[top=1in, left=1in, right=1in, bottom=1in]{geometry}
\usepackage{subfigure}
\usepackage{amssymb}
\usepackage{pifont}
\usepackage{mathtools}
\usepackage{stmaryrd}
\usepackage[T1]{fontenc}

\usepackage{authblk}
\usepackage{enumitem}
\usepackage{algorithmic}

\usepackage{tabularx}
\usepackage{caption}

\usepackage{wrapfig}

\theoremstyle{definition}

\renewcommand{\epsilon}{\varepsilon}
\renewcommand{\phi}{\varphi}

\makeatletter
\newcommand*{\RN}[1]{\expandafter\@slowromancap\romannumeral #1@}
\makeatother

\makeatletter
\newcommand{\printfnsymbol}[1]{%
  \textsuperscript{\@fnsymbol{#1}}%
}
\makeatother

\title{OAT-Rephrase: Optimization-Aware Training Data Rephrasing for Zeroth-Order LLM Fine-Tuning}

\author[1]{Jikai Long}
\author[ ]{Zijian Hu\thanks{Independent researcher}} 
\author[1]{Xiaodong Yu}
\author[2]{Jianwen Xie}
\author[1]{Zhaozhuo Xu\thanks{Corresponding author: \texttt{zxu79@stevens.edu}}}
\affil[1]{Stevens Institute of Technology}
\affil[2]{Lambda, Inc.}

\begin{document}

\maketitle
\begin{abstract}
    Fine-tuning large language models (LLMs) using zeroth-order optimization  (ZO) offers a memory-efficient alternative to gradient-based methods but suffers from slower convergence and unstable optimization due to noisy gradient estimates. This paper introduces OAT-Rephrase, an \underline{O}ptimization-\underline{A}ware \underline{T}raining data rephrasing strategy that leverages an LLM to rephrase training instances based on its understanding of the ZO dynamics, specifically MeZO, derived directly from its paper. The approach incorporates a dual-stage pipeline featuring a rewriter LLM and a semantic judge, ensuring all rephrasings retain task relevance and logical consistency. Evaluations across five classification tasks and three LLM architectures demonstrate that OAT-Rephrase consistently improves MeZO fine-tuning performance, often narrowing or eliminating the gap with first-order methods. Our findings suggest that optimization-aware rephrasing serves as a reusable and low-overhead enhancement for zeroth-order tuning regimes.
\end{abstract}

\section{Introduction}
Large language models (LLMs)~\cite{brown2020language,grattafiori2024llama,Mistral7b} have become central to natural language processing (NLP) but their adaptation to downstream tasks remains resource-intensive. Fine-tuning these models, particularly using gradient-based methods~\cite{liu2023winner,hu2022lora,liu2024boft}, requires substantial memory and compute due to the overhead of backpropagation. Zeroth-order optimization (ZO)~\cite{malladi2023fine,zhang2024zobench} has emerged as a promising alternative, offering memory-efficient fine-tuning by estimating gradients through forward passes alone. This approach is especially appealing for environments with limited computational resources, such as edge devices or single-GPU setups~\cite{guo2025zeroth}.

However, despite its advantages, ZO typically exhibits slower convergence and higher variance in gradient estimates~\cite{liu2018zeroth, malladi2023fine,guatam2024variance}, often leading to suboptimal model performance. Prior work has explored algorithmic adjustments, including tuning hyperparameters like perturbation magnitude and sampling strategies. However, these refinements have produced only limited improvements~\cite{yu2024subzero,zhang2024zobench}. This limitation suggests a need for complementary strategies that address ZO's sensitivity to data characteristics.

This paper presents a new perspective on adapting training data for ZO. Instead of modifying ZO methods to fit existing data, we explore how rephrasing training data can improve compatibility with ZO-based fine-tuning. Using MeZO~\cite{malladi2023fine} as a representative ZO method, we examine whether \textit{\textbf{a LLM can read and internalize the MeZO paper, then rewrite training instances to better align with its optimization process.}} To this end, we propose OAT-Rephrase, a structured workflow in which an LLM analyzes the MeZO paper and rewrites training examples in a way that preserves their original semantics while making them more suitable for MeZO-based fine-tuning.

Our contributions are:

\begin{itemize}[nosep,leftmargin=*]
    \item \textbf{Optimization-Aware Data Rephrasing:} We propose a pipeline that uses an LLM to read and interpret the MeZO method from its original paper, then rewrite training instances to better align with MeZO-based fine-tuning.

    \item \textbf{Rejection-Gated Rewriting Pipeline:} We introduce a rejection gate mechanism, supported by a separate LLM-based judge, to ensure that all rephrased training instances preserve the original semantics and logical consistency before being accepted into the synthetic corpus.

    \item \textbf{Generalizable Performance Gains Across Models:} We evaluate OAT-Rephrase on five classification tasks and three LLM backbones, demonstrating consistent improvements in MeZO-based fine-tuning—often matching or exceeding the performance of first-order methods.
\end{itemize}

\section{Problem Statement}
ZO for LLM fine-tuning~\cite{malladi2023fine,zhang2024zobench,guo2025zeroth} can be formulated as follows.
Let $\mathcal{T}$ represent the training corpus used to adapt a LLM with objective function $\mathcal{L}$. At each iteration, given model parameters $\boldsymbol{\theta} \in \mathbb{R}^n$, we randomly draw a minibatch $\mathcal{S} \subset \mathcal{T}$ and compute the update direction via:
\begin{align}\label{eq:zo_alt}
\widehat{\nabla} \mathcal{L}(\boldsymbol{\theta}; \mathcal{S}) = \boldsymbol{\xi} \cdot \frac{\mathcal{L}(\boldsymbol{\theta} + \delta \cdot \boldsymbol{\xi}; \mathcal{S})
- \mathcal{L}(\boldsymbol{\theta} - \delta \cdot \boldsymbol{\xi}; \mathcal{S})}
{2\delta}
\end{align}
where $\boldsymbol{\xi} \in \mathbb{R}^n$ is a stochastic direction vector with entries independently sampled from the standard Gaussian distribution $\mathcal{N}(0, 1)$, and $\delta \in \mathbb{R}$ denotes the perturbation magnitude. 

The resulting estimator $\widehat{\nabla} \mathcal{L}$ provides an approximation of the true gradient without requiring explicit backpropagation, which makes ZO particularly appealing in resource-constrained settings where LLM gradients are infeasible to compute. Next, ZO use the gradient estimated through Eq~\ref{eq:zo_alt} to update the parameter $\boldsymbol{\theta}$ with a learning rate.

\subsection{Pros and Cons of Zeroth-Order LLM Fine-Tuning}

\noindent \textbf{Pros: Memory-Efficient Fine-Tuning without Backpropagation.}
ZO eliminates the need for gradient computation through backpropagation, significantly reducing memory consumption during fine-tuning~\cite{malladi2023fine,guo2025zeroth}. This is particularly beneficial for LLMs, where storing intermediate activations during backpropagation can be prohibitively expensive on resource-constrained hardware such as edge devices or consumer GPUs~\cite{guo2025zeroth}. By estimating gradients through function evaluations rather than differentiation, ZO-based methods enables deployment in environments where first-order gradient-based training is infeasible. 

\noindent \textbf{Cons: Potential Performance Decline.}
ZO methods often suffer from higher variance and slower convergence compared to first-order approaches~\cite{malladi2023fine,zhang2024zobench}. In practice, they may lead to suboptimal performance relative to gradient-based methods like full fine-tuning and LoRA~\cite{hu2022lora}, or other parameter-efficient strategies~\cite{liu2023winner,ding2023parameter,liu2024boft,su2024defense}. Moreover, the reliance on stochastic direction sampling can introduce instability and requires careful tuning of the perturbation magnitude $\delta$.

\subsection{Optimization-Aware Training Data Rephrasing}

\noindent \textbf{Extensive Hyperparameter Tuning for ZO, But What Do We Get in Return?} 
Despite the appeal of ZO for memory-constrained fine-tuning, empirical studies~\cite{zhang2024zobench,guo2025zeroth} reveal that performance gains from extensive hyperparameter tuning, such as perturbation magnitude $\delta$, sampling strategy for $\boldsymbol{\xi}$, learning rate, and minibatch size, remain marginal and often inconsistent. This diminishing return highlights the need for complementary strategies beyond algorithmic tweaking to better align the optimization procedure with the data and task.

\noindent \textbf{Can We Rephrase Training Data to Better Support ZO?} 
Instead of solely focusing on adapting the hyperparameters associated with the ZO algorithm to the fixed data, we ask a complementary question: \textit{can the training data itself be restructured or rephrased to make ZO more effective? }Specifically, given that ZO methods optimize via finite-difference evaluations of the loss landscape, training data with noisy or ambiguous phrasing may introduce unnecessary variance in loss values, making the directional estimates less stable. Therefore, careful rephrasing of training inputs while preserving their original semantics may yield smoother and more optimizable loss surfaces for ZO methods to operate on. 

\begin{figure*}[!h]
    \centering
    \includegraphics[width=0.85\textwidth]{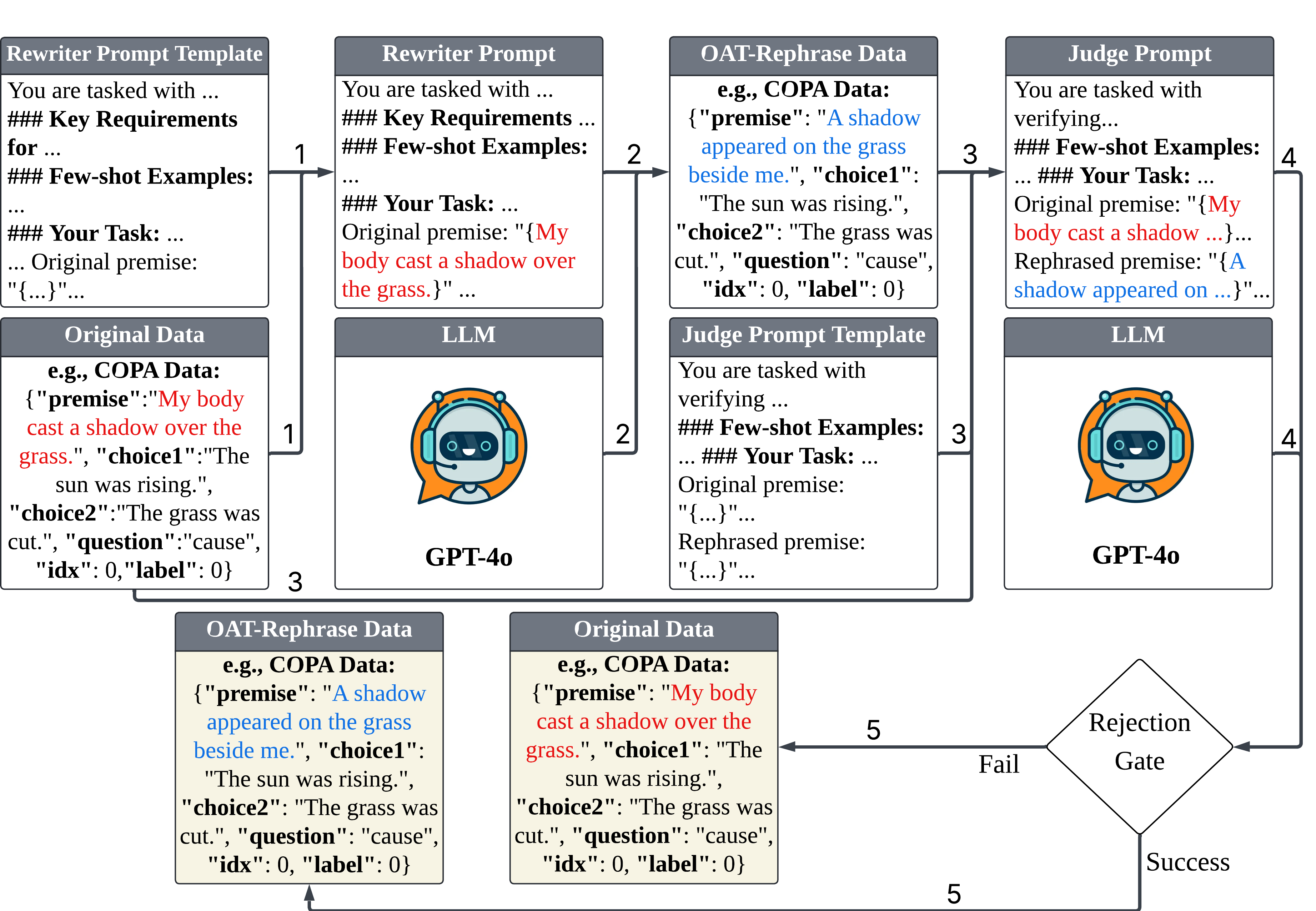}
    \caption{OAT-Rephrase workflow. (i) A \textbf{Rewriter Prompt} Template encodes task instructions, key requirements, and few-shot examples. Populated with an original instance (shown in red), it forms a \textbf{Rewriter Prompt}. (ii) An LLM generates a rewritten version (shown in blue). (iii) The original and rewritten content are composed into a \textbf{Judge Prompt} via a \textbf{Judge Prompt Template}. (iv) An LLM verifies whether the rewritten version preserves the original meaning and logical structure. The \textbf{Rejection Gate} discards unfaithful rewrites and retains the original; otherwise, the rewritten version is accepted. (v) This process repeats until the target corpus size is reached.}
    \label{fig:pipeline}
    \vspace{-6mm}
    \vskip 6pt
\end{figure*}

\noindent \textbf{Our Proposal: Let LLMs Understand ZO Dynamics and Rephrase Training Data for It Accordingly.}
We propose an optimization-aware data rephrasing pipeline in which another LLM is prompted to simulate the role of a ZO-aware training data rewriter. By being exposed to a ZO paper, the rewriter learns to rewrite training samples based on the its understanding of the paper. In essence, this makes the training corpus more ZO-friendly towards more stable convergence. Importantly, the rephrasing process is guided to retain original semantics and task relevance, ensuring that performance on downstream objectives remains unaffected. Our experiments demonstrate that this targeted rephrasing consistently improves ZO fine-tuning performance with minimal computational overhead.

\section{OAT-Rephrase Workflow to Generate Data Fit for ZO}
This section investigates whether an LLM, after having another LLM summarise the MeZO paper~\cite{malladi2023fine}, can rewrite an existing training dataset into a form that better suits ZO.  The workflow contains two components that are applied to each dataset in turn: an LLM based rewriter and an LLM based judge.  One language model produces a concise summary of the MeZO paper; that summary is placed in a prompt that guides a second language model, the rewriter, to generate a synthetic dataset by rephrasing the original training examples. The judge then checks every rephrased instance to confirm that it preserves both the semantic meaning and the logical consistency of its source. The OAT-Rephrase workflow is shown in Figure ~\ref{fig:pipeline}.

\subsection{An LLM-Based Rewriter for Training Data Rephrasing}

The rewrite for each dataset consists of an LLM (e.g., GPT-4o~\cite{hurst2024gpt}) combined with a prompt template. For each original training data instance in the dataset, we create a prompt using this template, which then guides the LLM to generate the rewritten version. As a result, the design of the prompt template is crucial to the rewriter’s effectiveness. In this section, we break down the prompt template design into five key phases.

\noindent \textbf{Phase 1: Constructing the Initial Prompt Template.} At the outset of prompt‑template design, we manually construct an initial template that contains four elements: (i) a task description (e.g., \emph{You are tasked with rephrasing the following premise while preserving its original meaning and ensuring consistency with its associated question and choices}); (ii) the verbatim abstract of the MeZO paper~\cite{malladi2023fine}; (iii) the verbatim introduction of the MeZO paper; (iv) textual paraphrases of the figures, tables, and Algorithm that appear within the introduction section of the MeZO paper; and (v) placeholder slots for a synthetic data instance. These components provide the LLM with sufficient contextual and methodological information about zeroth-order LLM fine-tuning for the subsequent template‑refinement stages. Using the COPA~\cite{roemmele2011copa} dataset as a representative case, we provide the prompt templates corresponding to each design phase in Appendix~\ref{sec:app_prompt}. The phase 1 template is shown in Figure~\ref{fig:phase1}.

\noindent \textbf{Phase 2: Fill MeZO Paper Information Gaps and Reaffirm the Goal of Prompt Optimization.} Because the initial prompt template contains only the abstract and introduction of the MeZO paper, we upload the full PDF of the MeZO paper to LLM to ensure that all relevant MeZO-related information is accurately retained. LLM is instructed to analyze the full document and compare it against the current prompt template, identifying any factual omissions or inaccuracies; if such issues are found, it is asked to revise the template accordingly. To further guide the rewriting process, we reiterate the objective of the prompt multiple times throughout the interaction: to generate rewritten data that is better suited for training with MeZO. Specifically, we aim for models trained on the synthetic data using MeZO to outperform those trained on the original data under the same optimization setting. Occasionally, after LLM has modified the prompt, we ask it to explain the reasoning behind a particular change it made. This is intended to help steer future revisions of the prompt template toward more targeted and effective improvements. Take the COPA dataset for example, the phase 2 template is shown in Figure~\ref{fig:phase2}.

\begin{wrapfigure}{r}{0.45\textwidth}  
\vspace{-6mm}
  \centering
  \includegraphics[width=0.86\linewidth]{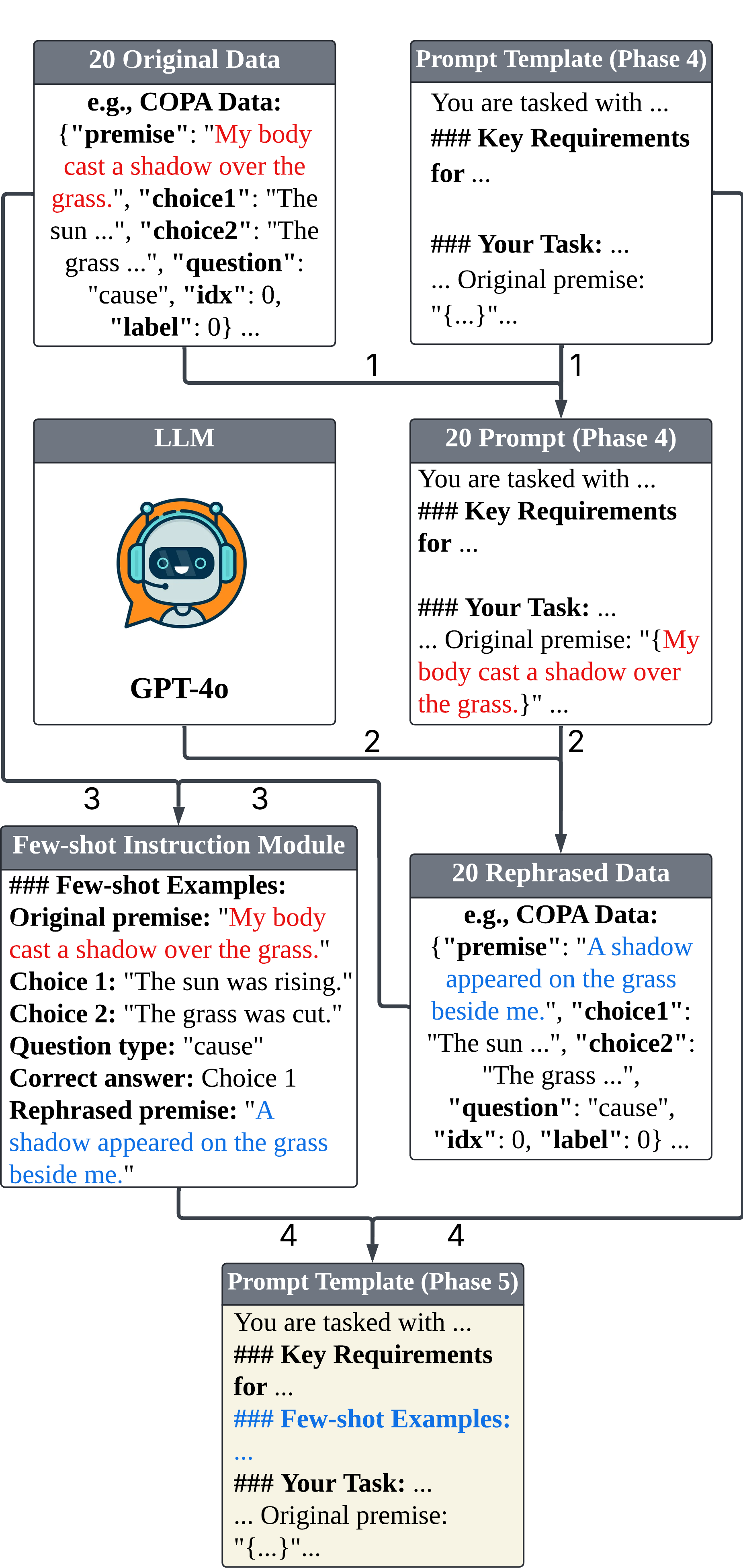}
  \captionsetup{width=0.45\textwidth} 
  \caption{Construction of the Few-Shot Instruction Module. (i) \textbf{20 Original Data} instances are inserted into the \textbf{Phase-4 Prompt Template} to generate \textbf{zero-shot rewriter prompts}, with the rewrite span marked in red. (ii) The LLM rewrites each span, yielding OAT-Rephrase outputs with rewritten text shown in blue. (iii) All rewritten results are manually inspected and paired with their original instance to form the \textbf{Few-Shot Instruction Module}. (iv) This module is appended to the prompt, resulting in a \textbf{Phase-5 Prompt Template} enriched with few-shot guidance.}
  \label{fig:rewriter}
  \vspace{-4mm}
\end{wrapfigure}

\noindent \textbf{Phase 3: Generation Error Correction.} The datasets targeted in this study consist of binary and multi-class classification tasks. For each data instance, the content subject to rewriting typically falls into one of two categories: the classification question itself and its accompanying background information. In our prompt template, we instruct the LLM to generate rewrites that preserve both the original sentence's meaning and its logical role within the data structure. However, this constraint frequently leads the LLM to inadvertently include the correct answer to the classification question in the rewritten sentence. To address this issue, we employ the prompt template as it stands at that point to rewrite 20 examples and manually examine whether any rewritten instances violate the requirements. If such violations are found, we provide LLM with the current template along with a representative faulty example and instruct it to revise the prompt. This process is repeated until 20 rewritten examples are deemed valid. The phase 3 COPA template is shown in Figure~\ref{fig:phase3}.

\noindent \textbf{Phase 4: MeZO Paper Information Augmentation.} During Phase 3, LLM made multiple adjustments to the prompt template based on input unrelated to MeZO, which led to a gradual loss of MeZO-specific information in the template. Therefore, it became necessary to reinforce the MeZO-related content within the prompt template. In this phase, we again provide the full MeZO paper to LLM and instruct it to analyze the original text and summarize key MeZO-related elements that should be preserved in the synthetic data. We then ask LLM to compare these key points against the current prompt template. If any essential information is missing, LLM is prompted to revise the template accordingly. After several rounds of comparison and refinement, and once all key MeZO information is fully captured, we make minor reorganizations and ordering adjustments, without removing any content, to produce the final version of the prompt template. The phase 4 COPA prompt template can be seen in Figure~\ref{fig:phase4}.

\begin{wrapfigure}{r}{0.45\textwidth}  
\vspace{-6mm}
  \centering
  \includegraphics[width=0.85\linewidth]{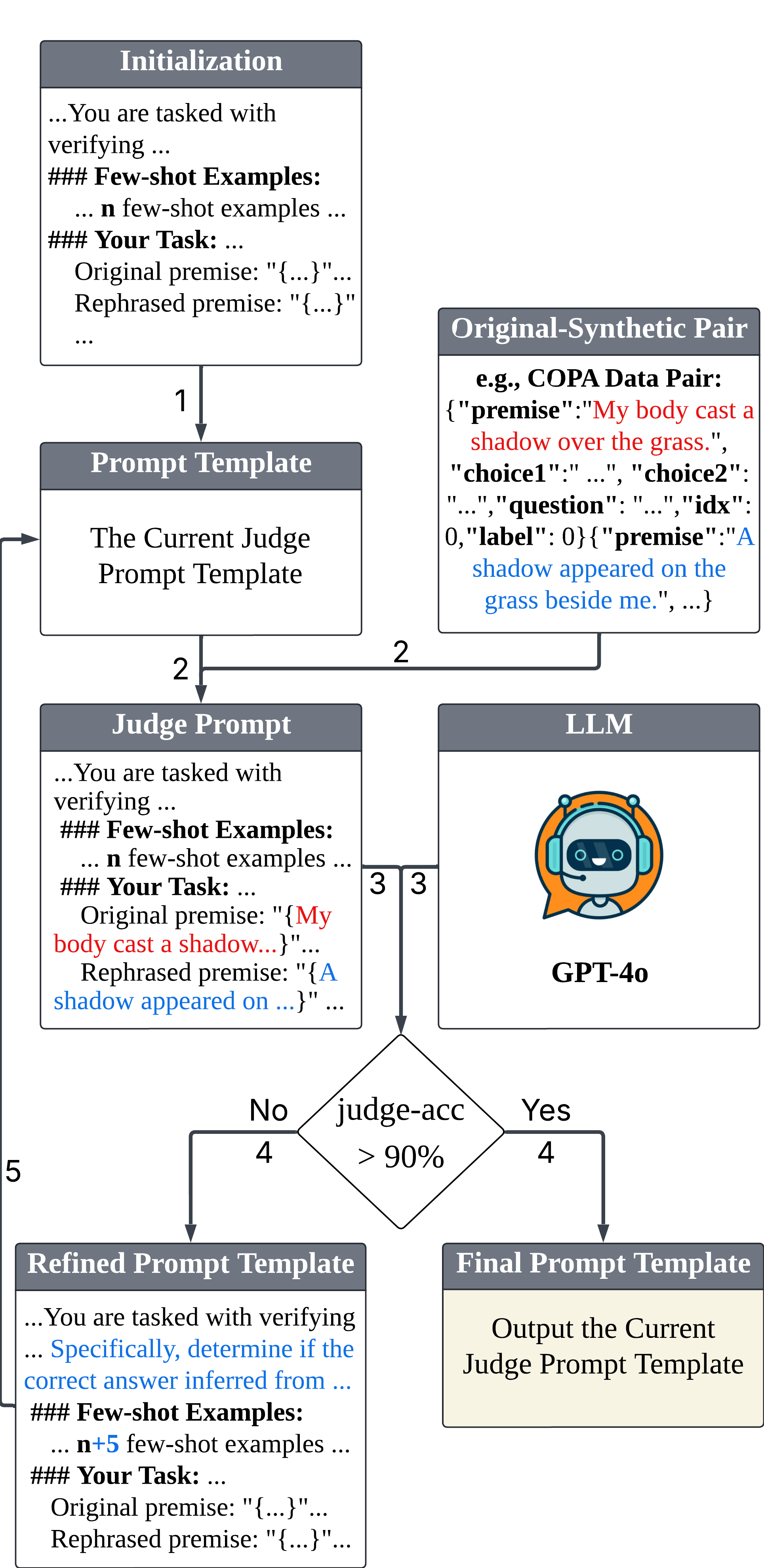}
  \captionsetup{width=0.45\textwidth} 
  \caption{\textbf{The Judge Prompt Template} starts with task instructions and five few-shot examples. In each refinement cycle, an \textbf{Original-Synthetic Pair} is added to form a \textbf{Judge Prompt}: spans copied from the original are shown in red, their rewrites in blue. The LLM evaluates forty prompts, and the resulting \textbf{judge accuracy}, defined as the proportion of model decisions that match human judgments, determines the process. A template that reaches 90 \% judge accuracy is fixed; otherwise, additional guidance and extra few-shot examples (blue additions in the Refined Prompt Template) are inserted, and the loop continues.}
  \label{fig:judge}
\vspace{-8mm}
\end{wrapfigure}

\noindent \textbf{Phase 5: Adding a Few-Shot Module.} To enable the rewriter to generate high-quality output reliably, the prompt template must be augmented with a sufficient number of few‑shot examples. We construct these few-shot examples through the following procedure. First, we apply the zero-shot version of the prompt template to rewrite 20 training data instances. Then, each rewritten instance is manually compared against its original counterpart and evaluated according to our quality standards. Finally, the instances that pass this evaluation are paired with their originals and reformatted into a dedicated schema (see Figure~\ref{fig:rewriter} for the COPA dataset), which is subsequently inserted into the prompt template as few-shot examples. See the phase 5 prompt template in Figure~\ref{fig:phase5}.

\noindent \textbf{Template Transfer.} The rewriter prompt template can be ported to fit specific dataset with only light editing.  First, replace the template’s schema description with the field names and logical roles used by the target dataset.  Next, remove the existing Few-Shot Module and re-run Phase 5 to create a new Few-Shot Module.  This two-step adjustment aligns the template with the new data format and restores the guidance required for high-quality rewriting, completing the transfer.

\subsection{An LLM-Based Judge for Preserving Meaning in Training Data}
To qualify as acceptable, each rewritten training data instance must faithfully preserve the original sentence’s meaning and maintain the same logical role within the overall data structure.  This requirement prevents label corruption, avoids task drift, and ensures that performance differences between models trained on synthetic versus original data reflect genuine improvements rather than artifacts of semantic divergence.  To enforce this standard, we introduce an LLM-based judge that compares each synthetic–original pair and rejects rewrite that violates semantic or logical consistency.

\noindent \textbf{Judge Design.} Each dataset‑specific judge mirrors the architecture of the rewriter: it consists of an LLM (GPT‑4o) and a dedicated prompt template. For every original–synthetic pair, the template embeds both instances in a single prompt and asks the LLM to decide (i) whether the rewritten span preserves the meaning of the corresponding span in the original sentence and (ii) whether the overall logical relationship in the instance remains unchanged. After this judging step, a rejection gate is applied: any synthetic instance that fails either criterion is discarded, and its original counterpart is reinstated in its place within the synthetic dataset. Only examples that pass both checks are retained as part of the rephrased synthetic dataset.

\noindent \textbf{Prompt Template Construction and Iterative Refinement.} We build and tune the Judge Prompt Template using the following procedure (shown in Figure~\ref{fig:judge} ): (i) Initialization: we manually design an initial prompt template (see Figure~\ref{fig:judge1} for the COPA dataset) that contains three elements: a task description; five hand‑labelled success cases that serve as few‑shot examples; and reserve slots for the original–synthetic pair on which the LLM must output \emph{same} or \emph{not the same}. (ii) Evaluation loop: we build a forty-pair test set whose human labels indicate whether each rephrased premise remains semantically and logically consistent with its original. Passing these pairs through the judge yields a metric, hereafter called \textbf{judge accuracy} (judge-acc), defined as the proportion of model decisions that match the human annotations. If judge-acc reaches at least 90\%, the prompt template is deemed reliable and used to filter the full synthetic corpus; otherwise, the template is revised and the evaluation is repeated. (iii) Revision strategies: when judge-acc is below 90\%, we refine the task description by adding more detailed, dataset-specific requirements (e.g., see the task description for the COPA dataset in the final version of the Judge Prompt Template shown in Figure~\ref{fig:judge2}), and we enlarge the few-shot set (up to twenty examples for COPA). Steps (ii) and (iii) are repeated until the template achieves the 90\% threshold, after which the finalised template is employed for large-scale filtering of the synthetic corpus.

\section{Experiment}
We perform evaluation for OAT-Rephrase to answer the following research questions.
\begin{itemize}[nosep,leftmargin=*]
    \item \textbf{RQ1:} Does fine-tuning an LLM with MeZO on OAT-Rephrase data consistently outperform fine-tuning with MeZO on the original data?
    \item \textbf{RQ2:} Can OAT-Rephrase enable a MeZO-tuned LLM to surpass the performance of the same LLM fine-tuned with first-order optimization methods such as OAT-Rephrase?
    \item \textbf{RQ3:} If rephrased data improves MeZO-based LLM fine-tuning, does this benefit extend across various LLM architectures?
\end{itemize}

\subsection{Experiment Setup} 
We benchmark OAT-Rephrase on five English classification tasks and three open-source LLM backbones. For every model-dataset pair we evaluate six training settings; all runs set the random seed as 0 for our method and baselines to make sure the noise vector remains the same.

\noindent \textbf{Datasets.} We study four SuperGLUE tasks (COPA~\cite{roemmele2011copa}, CB~\cite{de2019cb}, RTE~\cite{wang2018glue}, and BoolQ~\cite{clark2019boolq}) together with ARC-Challenge (ARC-C)~\cite{clark2018arc}. We perform train/dev/test split following ~\cite{guo2025zeroth}. The identical data indices are used for both the Original and OAT-Rephrase variants.

\begin{table}[!h]
\centering
\caption{Accuracy of the rewriter on each dataset, as assessed by the LLM-based judge. }
\label{tab:rewriter-acc}
\begin{tabularx}{\textwidth}{lXXXXX}
\toprule
& \textbf{COPA} & \textbf{CB} & \textbf{RTE} & \textbf{BoolQ} & \textbf{ARC-C} \\
\midrule
\textbf{Rewriter Acc} & 100.0 & 99.20 & 100.0 & 96.20 & 98.03 \\
\bottomrule
\end{tabularx}
\end{table}
\noindent\textbf{Rewriter Accuracy.}  
We define \textbf{rewriter accuracy} as the proportion of rewritten training items that our LLM-based judge considers both semantically and logically consistent with their originals. A rewritten corpus is accepted only when its rewriter accuracy reaches at least 90 \%. The COPA rewriter prompt template is produced through the full five-phase design, whereas the rewriter prompt templates for CB, RTE, BoolQ, and ArcC are obtained via the template transfer process from COPA. All transferred templates reach judge accuracies between 96.2 \% and 100.0 \% (see Table \ref{tab:rewriter-acc}), so we retain them without additional revision in the subsequent experiments.

\noindent \textbf{LLM Training Setting.}
For every model-dataset pair, we conduct 6 training conditions, each combining one of 3 optimization regimes with either the Original or the OAT-Rephrase training slice:
\begin{itemize}[nosep, leftmargin=*]
    \item ZO-MeZO on Original
    \item ZO-MeZO on OAT-Rephrase (for MeZO)
    \item AdamW, full fine-tuning (FO-Full) on Original
    \item AdamW, FO-Full on OAT-Rephrase (for MeZO)
    \item AdamW, LoRA fine-tuning (FO-LoRA)~\cite{hu2022lora} on Original
    \item AdamW, FO-LoRA on OAT-Rephrase (for MeZO)
\end{itemize}
This design isolates the influence of rewritten data across three distinct training methods, namely MeZO optimization, first-order full fine-tuning, and first-order LoRA fine-tuning, while keeping all other factors constant.

\begin{table*}[h!]
\centering
\caption{Test accuracy of LLM backbones fine-tuned on either the \textbf{Original} or the \textbf{OAT-Rephrase} version of the same training split. Within each sub-table, \textbf{Data Type} indicates the dataset variant, \textbf{Methods} specifies the optimisation regime (MeZO, first-order full-parameter fine-tuning, or first-order LoRA fine-tuning), and the task columns report accuracy on five benchmarks: COPA, CB, RTE, BoolQ, and ARC-C. \textbf{Avg. Acc.} refers to the average accuracy across all five tasks. \textbf{Avg.\,$\Delta$} denotes the mean difference between OAT-Rephrase and Original; a positive value indicates that rephrasing the training data improves performance.}
\label{tab:main-results}
\small
\setlength{\tabcolsep}{4pt}  
\renewcommand{\arraystretch}{1.0}

\begin{tabular*}{1\linewidth}{@{\extracolsep{\fill}}l l ccccccc@{}}
\multicolumn{9}{c}{\textbf{(a) \,Llama-3.2-1B}}\\[2pt]
\toprule
\textbf{Data Type} & \textbf{Methods} & COPA & CB & RTE & BoolQ & ARC-C & \textbf{Avg. Acc.} & \textbf{Avg.\,$\Delta$}\\
\midrule
Original     & \multirow{2}{*}{ZO-MeZO} & 77.0\textsubscript{} & 66.1\textsubscript{} & 58.5\textsubscript{} & 64.1\textsubscript{} & 36.1\textsubscript{} & 60.36\textsubscript{}\\
OAT-Rephrase &                               & \underline{\textbf{79.0}}\textsubscript{} & \textbf{71.4}\textsubscript{} & \underline{\textbf{61.4}}\textsubscript{} & \textbf{64.3}\textsubscript{} & \textbf{37.1}\textsubscript{} & \textbf{62.64}\textsubscript{} & \textbf{2.28}\textsubscript{}\\
\cmidrule(l){1-9}
Original     & \multirow{2}{*}{FO-Full} & 78.0\textsubscript{} & 66.1\textsubscript{} & 54.9\textsubscript{} & 70.6\textsubscript{} & 43.1\textsubscript{} & 62.54\textsubscript{}\\
OAT-Rephrase &                               & 76.0\textsubscript{} & 66.1\textsubscript{} & 56.3\textsubscript{} & 71.3\textsubscript{} & 42.1\textsubscript{} & 62.36\textsubscript{} & -0.18\textsubscript{}\\
\cmidrule(l){1-9}
Original     & \multirow{2}{*}{FO-LoRA}  & 76.0\textsubscript{} & 67.9\textsubscript{} & 54.9\textsubscript{} & 72.3\textsubscript{} & 43.5\textsubscript{} & 62.92\textsubscript{}\\
OAT-Rephrase &                               & 77.0\textsubscript{} & 73.2\textsubscript{} & 57.8\textsubscript{} & 73.5\textsubscript{} & 43.1\textsubscript{} & 64.92\textsubscript{} & 2.00\textsubscript{}\\
\bottomrule
\end{tabular*}

\begin{tabular*}{1\linewidth}{@{\extracolsep{\fill}}ll ccccccc@{}}
\multicolumn{9}{c}{\textbf{(b) \,Llama-3.2-3B}}\\[2pt]
\toprule
\textbf{Data Type} & \textbf{Methods} & COPA & CB & RTE & BoolQ & ARC-C & \textbf{Avg. Acc.} & \textbf{Avg.\,$\Delta$}\\
\midrule
Original     & \multirow{2}{*}{ZO-MeZO} & 84.0\textsubscript{} & 69.6\textsubscript{} & 66.4\textsubscript{} & 74.9\textsubscript{} & 46.2\textsubscript{} & 68.22\textsubscript{}\\
OAT-Rephrase &                               & \textbf{86.0}\textsubscript{} & \underline{\textbf{71.4}}\textsubscript{} & 61.7\textsubscript{} & \textbf{77.2}\textsubscript{} & \textbf{47.8}\textsubscript{} & \textbf{68.82}\textsubscript{} & \textbf{0.60}\textsubscript{}\\
\cmidrule(l){1-9}
Original     & \multirow{2}{*}{FO-Full} & 86.0\textsubscript{} & 58.9\textsubscript{} & 75.1\textsubscript{} & 79.8\textsubscript{} & 50.5\textsubscript{} & 70.06\textsubscript{}\\
OAT-Rephrase &                               & 88.0\textsubscript{} & 50.0\textsubscript{} & 70.4\textsubscript{} & 80.7\textsubscript{} & 50.5\textsubscript{} & 67.92\textsubscript{} & -2.14\textsubscript{}\\
\cmidrule(l){1-9}
Original     & \multirow{2}{*}{FO-LoRA}  & 89.0\textsubscript{} & 64.3\textsubscript{} & 79.4\textsubscript{} & 81.6\textsubscript{} & 55.9\textsubscript{} & 74.04\textsubscript{}\\
OAT-Rephrase &                               & 89.0\textsubscript{} & 53.6\textsubscript{} & 71.80\textsubscript{} & 81.6\textsubscript{} & 54.2\textsubscript{} & 70.04\textsubscript{} & -4.00\textsubscript{}\\
\bottomrule
\end{tabular*}

\begin{tabular*}{1\linewidth}{@{\extracolsep{\fill}}ll ccccccc@{}}
\multicolumn{9}{c}{\textbf{(c) \,Mistral-7B-v0.1}}\\[2pt]
\toprule
\textbf{Data Type} & \textbf{Methods} & COPA & CB & RTE & BoolQ & ARC-C & \textbf{Avg. Acc.} & \textbf{Avg.\,$\Delta$}\\
\midrule
Original     & \multirow{2}{*}{ZO-MeZO} & 59.0\textsubscript{} & 50.0\textsubscript{} & 51.6\textsubscript{} & 63.1\textsubscript{} & 48.2\textsubscript{} & 54.38\textsubscript{}\\
OAT-Rephrase &                               & \textbf{64.0}\textsubscript{} & \textbf{53.6}\textsubscript{} & \textbf{54.2}\textsubscript{} & \textbf{63.2}\textsubscript{} & \textbf{48.5}\textsubscript{} & \textbf{56.70}\textsubscript{} & \textbf{2.32}\textsubscript{}\\
\cmidrule(l){1-9}
Original     & \multirow{2}{*}{FO-Full} & 90.0\textsubscript{} & 67.9\textsubscript{} & 74.7\textsubscript{} & 80.2\textsubscript{} & 61.9\textsubscript{} & 74.94\textsubscript{}\\
OAT-Rephrase &                               & 91.0\textsubscript{} & 64.3\textsubscript{} & 77.3\textsubscript{} & 79.5\textsubscript{} & 67.6\textsubscript{} & 75.94\textsubscript{} & 1.00\textsubscript{}\\
\cmidrule(l){1-9}
Original     & \multirow{2}{*}{FO-LoRA}  & 90.0\textsubscript{} & 75.0\textsubscript{} & 79.8\textsubscript{} & 86.3\textsubscript{} & 71.9\textsubscript{} & 80.60\textsubscript{}\\
OAT-Rephrase &                               & 91.0\textsubscript{} & 69.6\textsubscript{} & 79.1\textsubscript{} & 84.8\textsubscript{} & 71.6\textsubscript{} & 79.22\textsubscript{} & -1.38\textsubscript{}\\
\bottomrule
\end{tabular*}
\end{table*}

\noindent \textbf{Models and Hyperparameters.}  
We fine-tune three open-source checkpoints: Llama 3.2-1B, Llama 3.2-3B and Mistral-7B-v0.1. All experiments use bfloat16 precision and run on either one or two NVIDIA A100 GPUs with 40 GB memory, one or two NVIDIA A6000 GPUs with 48 GB memory, or a single NVIDIA GH200 GPU with 96 GB memory. Each backbone is trained with a fixed token budget to ensure comparable compute: 

\begin{itemize}[nosep,leftmargin=*]

    \item \textbf{Llama-3.2-1B/3B}~\cite{grattafiori2024llama} is trained with batch size 16 for 20 k update steps. When GPU memory is insufficient, we adopt alternative configurations that preserve the total number of processed tokens: either halving the batch size and doubling the number of update steps (e.g., 8/40k, 4/80k or 1/320k), or using a batch size of 4 with gradient accumulation of 4. 
    \item \textbf{Mistral-7B-v0.1}~\cite{Mistral7b} is trained with a batch size of 4 for 80k steps; When memory constraints arise, we fall back to either (i) a batch size of 2 for 160 k steps, (ii) a batch size of 1 for 320 k steps, or (iii) a batch size of 1 for 80 k steps with a gradient-accumulation of 4.

\end{itemize}

A global random seed of 0 fixes data order, dropout masks and parameter initialisation across every experiment.
A unified grid prevents any method from benefiting from hand-picked settings. \textbf{ZO-MeZO} and \textbf{full fine-tuning} sweep
learning rates $lr$ $\!\in\!\{1\!\times\!10^{-6},\;5\!\times\!10^{-7},\;2\!\times\!10^{-7},\;1\!\times\!10^{-7}\}$  
with batch sizes $bs$ $\!\in\!\{2,4,8,16\}$; all MeZO runs fix $\delta = 10^{-3}$ ($\delta$ in Eq~\ref{eq:zo_alt}). \noindent \textbf{LoRA fine-tuning} sweeps ranks $rank\!\in\!\{8,16,32\}$ and
$lr\!\in\!\{1\!\times\!10^{-4},\;2\!\times\!10^{-4}\}$.  
For each run we select the configuration that maximises development accuracy and report its test result (as shown in Table ~\ref{tab:main-results}).

\noindent \textbf{Evaluation Metric.}
We report classification accuracy on the held-out Test set.  Because the same example indices, seed and optimiser noise are used across paired runs, any performance difference reflects the training method or the rewritten data, not stochastic variance.

\subsection{Answers to RQ1: OAT-Rephrase Improves the MeZO Fine-Tuning}

As shown in Table~\ref{tab:main-results}, OAT-Rephrase consistently improves MeZO-based fine-tuning across nearly all tasks and model scales. COPA and CB show the most stable gains across models, each exceeding 2 points. Notably, CB improves by 5.3 points on Llama-3.2-1B, and COPA by 5.0 points on Mistral-7B-v0.1. RTE also improves on Llama-3.2-1B and Mistral-7B-v0.1, while BoolQ gains 2.3 points on Llama-3.2-3B. Every task except RTE improves on all models. These task-level gains result in average accuracy increases of 2.28, 0.60, and 2.32 on Llama-3.2-1B, Llama-3.2-3B, and Mistral-7B-v0.1. This suggests OAT-Rephrase provides clearer and more MeZO-compatible formulations of the original data, enabling MeZO to produce better updates from forward-only feedback.

\subsection{Answers to RQ2: OAT-Rephrase Uplift MeZO to the Level of First-Order Methods}

OAT-Rephrase effectively narrows the gap between MeZO and first-order methods. In some settings, ZO-MeZO with OAT-Rephrase outperforms FO-Full or FO-LoRA using Original data. For instance, on CB with Llama-3.2-1B, MeZO achieves 71.4 accuracy compared to FO-Full’s 66.1. In other cases, MeZO nearly matches the first-order baseline, such as BoolQ on Llama-3.2-3B where MeZO reaches 77.2 and FO-Full reaches 79.8. Even where gaps persist, MeZO with OAT-Rephrase improves reliably across tasks and models, thereby reducing its historical disadvantage. These outcomes show that OAT-Rephrase substantially enhances MeZO’s competitiveness relative to first-order fine-tuning (Table~\ref{tab:main-results}).

\subsection{Answers to RQ3: One-Time OAT-Rephrase Generalizes to Multiple LLMs}

The rewritten datasets are generated once per task and reused across Llama-3.2-1B, 3B, and Mistral-7B-v0.1, consistently improving MeZO performance. This consistent benefit all the models indicates that OAT-Rephrase encodes task semantics in a way that generalizes beyond the model that produced the synthetic data. In contrast, as shown in Table~\ref{tab:main-results}, FO-Full improves only on Mistral-7B-v0.1 and FO-LoRA only on Llama-3.2-1B, with results on other models either plateauing or declining. This contrast highlights that OAT-Rephrase is particularly effective as a data augmentation strategy for MeZO, while offering less reliable gains for other optimization methods.

\section{Related Works}
\subsection{Zeorth-Order LLM Fine-Tuning}
Recent work has investigated ZO as a memory-efficient alternative to gradient-based fine-tuning. \citet{malladi2023fine} introduced MeZO, an in-place ZO method that requires only forward passes, significantly reducing memory overhead compared to backpropagation. \citet{zhang2024zobench} conducted a comprehensive evaluation of ZO techniques, noting their compatibility with both full-parameter and parameter-efficient tuning strategies like LoRA. Despite these advances, ZO methods still lag behind in performance due to high variance in gradient estimation. \citet{guo2025zeroth} proposed incorporating sparsity constraints to improve ZO stability, yet these improvements remain sensitive to hyperparameter choices.

\subsection{Training Data Synthesis for LLM}n 
Recent advances in synthetic data generation using LLMs have transformed data-centric strategies in NLP, enabling efficient augmentation and rewriting of training data to improve model performance. As outlined by~\cite{long-etal-2024-llms, he2022generate, ding2024data}, LLMs can generate diverse, high-quality data through carefully designed prompts and multi-step workflows. 
 These strategies enhance training effectiveness, particularly in low-resource or domain-specific settings~\cite{zhou2024survey}, positioning synthetic data as a powerful complement to traditional data collection and optimization methods.

\section{Conclusion}
This work rethinks the adaptation of training data for ZO LLM fine-tuning by introducing OAT-Rephrase, a structured pipeline that uses LLMs to rephrase training data with sensitivity to ZO dynamics. Unlike prior approaches that solely refine ZO algorithms or hyperparameters, OAT-Rephrase shifts the focus to data preprocessing, enabling LLMs to generate semantically equivalent, optimization-aligned examples based on a reading of the ZO method (MeZO). Our experiments across multiple datasets and model sizes show that this synthetic rephrasing significantly boosts MeZO’s effectiveness, often rivaling full gradient-based fine-tuning. Furthermore, the synthetic data generalizes well across models and optimization methods, making OAT-Rephrase a flexible tool for low-resource fine-tuning scenarios.

\bibliographystyle{plainnat}
\bibliography{ref}

\begin{thebibliography}{24}
\providecommand{\natexlab}[1]{#1}
\providecommand{\url}[1]{\texttt{#1}}
\expandafter\ifx\csname urlstyle\endcsname\relax
  \providecommand{\doi}[1]{doi: #1}\else
  \providecommand{\doi}{doi: \begingroup \urlstyle{rm}\Url}\fi

\bibitem[Brown et~al.(2020)Brown, Mann, Ryder, Subbiah, Kaplan, Dhariwal, Neelakantan, Shyam, Sastry, Askell, et~al.]{brown2020language}
Tom Brown, Benjamin Mann, Nick Ryder, Melanie Subbiah, Jared~D Kaplan, Prafulla Dhariwal, Arvind Neelakantan, Pranav Shyam, Girish Sastry, Amanda Askell, et~al.
\newblock Language models are few-shot learners.
\newblock \emph{Advances in neural information processing systems}, 33:\penalty0 1877--1901, 2020.

\bibitem[Clark et~al.(2019)Clark, Gardner, and Others]{clark2019boolq}
Christopher Clark, Mark Gardner, and Others.
\newblock Boolq: Exploring the surprising difficulty of natural yes/no questions.
\newblock In \emph{Proceedings of NAACL-HLT}, 2019.

\bibitem[Clark et~al.(2018)Clark, Cowhey, Etzioni, Khot, Sabharwal, Schoenick, and Tafjord]{clark2018arc}
Peter Clark, Isaac Cowhey, Oren Etzioni, Tushar Khot, Ashish Sabharwal, Carissa Schoenick, and Oyvind Tafjord.
\newblock Think you have solved question answering? try arc, the ai2 reasoning challenge.
\newblock In \emph{arXiv preprint arXiv:1803.05457}, 2018.

\bibitem[De~Marneffe et~al.(2019)De~Marneffe, Simons, and Tonhauser]{de2019cb}
Marie-Catherine De~Marneffe, Mandy Simons, and Judith Tonhauser.
\newblock The commitmentbank: Investigating projection in naturally occurring discourse.
\newblock In \emph{Proceedings of Sinn und Bedeutung}, 2019.

\bibitem[Ding et~al.(2024)Ding, Qin, Zhao, Luo, Li, Chen, Xia, Hu, Tuan, and Joty]{ding2024data}
Bosheng Ding, Chengwei Qin, Ruochen Zhao, Tianze Luo, Xinze Li, Guizhen Chen, Wenhan Xia, Junjie Hu, Luu~Anh Tuan, and Shafiq Joty.
\newblock Data augmentation using llms: Data perspectives, learning paradigms and challenges.
\newblock In \emph{Findings of the Association for Computational Linguistics ACL 2024}, pages 1679--1705, 2024.

\bibitem[Ding et~al.(2023)Ding, Qin, Yang, Wei, Yang, Su, Hu, Chen, Chan, Chen, et~al.]{ding2023parameter}
Ning Ding, Yujia Qin, Guang Yang, Fuchao Wei, Zonghan Yang, Yusheng Su, Shengding Hu, Yulin Chen, Chi-Min Chan, Weize Chen, et~al.
\newblock Parameter-efficient fine-tuning of large-scale pre-trained language models.
\newblock \emph{Nature Machine Intelligence}, 5\penalty0 (3):\penalty0 220--235, 2023.

\bibitem[Gautam et~al.(2024)Gautam, Park, Zhou, Raman, and Ha]{guatam2024variance}
Tanmay Gautam, Youngsuk Park, Hao Zhou, Parameswaran Raman, and Wooseok Ha.
\newblock Variance-reduced zeroth-order methods for fine-tuning language models.
\newblock In \emph{Forty-first International Conference on Machine Learning, {ICML} 2024, Vienna, Austria, July 21-27, 2024}. OpenReview.net, 2024.
\newblock URL \url{https://openreview.net/forum?id=VHO4nE7v41}.

\bibitem[Grattafiori et~al.(2024)Grattafiori, Dubey, Jauhri, Pandey, Kadian, Al-Dahle, Letman, Mathur, Schelten, Vaughan, et~al.]{grattafiori2024llama}
Aaron Grattafiori, Abhimanyu Dubey, Abhinav Jauhri, Abhinav Pandey, Abhishek Kadian, Ahmad Al-Dahle, Aiesha Letman, Akhil Mathur, Alan Schelten, Alex Vaughan, et~al.
\newblock The llama 3 herd of models.
\newblock \emph{arXiv preprint arXiv:2407.21783}, 2024.

\bibitem[Guo et~al.(2025)Guo, Long, Zeng, Liu, Yang, Ran, Gardner, Bastani, Sa, Yu, Chen, and Xu]{guo2025zeroth}
Wentao Guo, Jikai Long, Yimeng Zeng, Zirui Liu, Xinyu Yang, Yide Ran, Jacob~R. Gardner, Osbert Bastani, Christopher~De Sa, Xiaodong Yu, Beidi Chen, and Zhaozhuo Xu.
\newblock Zeroth-order fine-tuning of llms with transferable static sparsity.
\newblock In \emph{The Thirteenth International Conference on Learning Representations, {ICLR} 2025, Singapore, April 24-28, 2025}. OpenReview.net, 2025.
\newblock URL \url{https://openreview.net/forum?id=myYzr50xBh}.

\bibitem[He et~al.(2022)He, Nassar, Kiros, Haffari, and Norouzi]{he2022generate}
Xuanli He, Islam Nassar, Jamie Kiros, Gholamreza Haffari, and Mohammad Norouzi.
\newblock Generate, annotate, and learn: Nlp with synthetic text.
\newblock \emph{Transactions of the Association for Computational Linguistics}, 10:\penalty0 826--842, 2022.

\bibitem[Hu et~al.(2022)Hu, Shen, Wallis, Allen{-}Zhu, Li, Wang, Wang, and Chen]{hu2022lora}
Edward~J. Hu, Yelong Shen, Phillip Wallis, Zeyuan Allen{-}Zhu, Yuanzhi Li, Shean Wang, Lu~Wang, and Weizhu Chen.
\newblock Lora: Low-rank adaptation of large language models.
\newblock In \emph{The Tenth International Conference on Learning Representations, {ICLR} 2022, Virtual Event, April 25-29, 2022}. OpenReview.net, 2022.
\newblock URL \url{https://openreview.net/forum?id=nZeVKeeFYf9}.

\bibitem[Hurst et~al.(2024)Hurst, Lerer, Goucher, Perelman, Ramesh, Clark, Ostrow, Welihinda, Hayes, Radford, et~al.]{hurst2024gpt}
Aaron Hurst, Adam Lerer, Adam~P Goucher, Adam Perelman, Aditya Ramesh, Aidan Clark, AJ~Ostrow, Akila Welihinda, Alan Hayes, Alec Radford, et~al.
\newblock Gpt-4o system card.
\newblock \emph{arXiv preprint arXiv:2410.21276}, 2024.

\bibitem[Jiang et~al.(2023)Jiang, Sablayrolles, Mensch, Bamford, Chaplot, de~Las~Casas, Bressand, Lengyel, Lample, Saulnier, Lavaud, Lachaux, Stock, Scao, Lavril, Wang, Lacroix, and Sayed]{Mistral7b}
Albert~Q. Jiang, Alexandre Sablayrolles, Arthur Mensch, Chris Bamford, Devendra~Singh Chaplot, Diego de~Las~Casas, Florian Bressand, Gianna Lengyel, Guillaume Lample, Lucile Saulnier, L{\'{e}}lio~Renard Lavaud, Marie{-}Anne Lachaux, Pierre Stock, Teven~Le Scao, Thibaut Lavril, Thomas Wang, Timoth{\'{e}}e Lacroix, and William~El Sayed.
\newblock Mistral 7b.
\newblock \emph{CoRR}, abs/2310.06825, 2023.
\newblock \doi{10.48550/ARXIV.2310.06825}.
\newblock URL \url{https://doi.org/10.48550/arXiv.2310.06825}.

\bibitem[Liu et~al.(2018)Liu, Kailkhura, Chen, Ting, Chang, and Amini]{liu2018zeroth}
Sijia Liu, Bhavya Kailkhura, Pin-Yu Chen, Paishun Ting, Shiyu Chang, and Lisa Amini.
\newblock Zeroth-order stochastic variance reduction for nonconvex optimization.
\newblock \emph{Advances in neural information processing systems}, 31, 2018.

\bibitem[Liu et~al.(2024)Liu, Qiu, Feng, Xiu, Xue, Yu, Feng, Liu, Heo, Peng, Wen, Black, Weller, and Sch{\"{o}}lkopf]{liu2024boft}
Weiyang Liu, Zeju Qiu, Yao Feng, Yuliang Xiu, Yuxuan Xue, Longhui Yu, Haiwen Feng, Zhen Liu, Juyeon Heo, Songyou Peng, Yandong Wen, Michael~J. Black, Adrian Weller, and Bernhard Sch{\"{o}}lkopf.
\newblock Parameter-efficient orthogonal finetuning via butterfly factorization.
\newblock In \emph{The Twelfth International Conference on Learning Representations, {ICLR} 2024, Vienna, Austria, May 7-11, 2024}. OpenReview.net, 2024.
\newblock URL \url{https://openreview.net/forum?id=7NzgkEdGyr}.

\bibitem[Liu et~al.(2023)Liu, Wang, Zhong, Xu, Zha, Tang, Jiang, Zhou, Chaudhary, Xu, et~al.]{liu2023winner}
Zirui Liu, Guanchu Wang, Shaochen~Henry Zhong, Zhaozhuo Xu, Daochen Zha, Ruixiang~Ryan Tang, Zhimeng~Stephen Jiang, Kaixiong Zhou, Vipin Chaudhary, Shuai Xu, et~al.
\newblock Winner-take-all column row sampling for memory efficient adaptation of language model.
\newblock \emph{Advances in Neural Information Processing Systems}, 36:\penalty0 3402--3424, 2023.

\bibitem[Long et~al.(2024)Long, Wang, Xiao, Zhao, Ding, Chen, and Wang]{long-etal-2024-llms}
Lin Long, Rui Wang, Ruixuan Xiao, Junbo Zhao, Xiao Ding, Gang Chen, and Haobo Wang.
\newblock On {LLM}s-driven synthetic data generation, curation, and evaluation: A survey.
\newblock In Lun-Wei Ku, Andre Martins, and Vivek Srikumar, editors, \emph{Findings of the Association for Computational Linguistics: ACL 2024}, pages 11065--11082, Bangkok, Thailand, August 2024. Association for Computational Linguistics.
\newblock \doi{10.18653/v1/2024.findings-acl.658}.
\newblock URL \url{https://aclanthology.org/2024.findings-acl.658/}.

\bibitem[Malladi et~al.(2023)Malladi, Gao, Nichani, Damian, Lee, Chen, and Arora]{malladi2023fine}
Sadhika Malladi, Tianyu Gao, Eshaan Nichani, Alex Damian, Jason~D Lee, Danqi Chen, and Sanjeev Arora.
\newblock Fine-tuning language models with just forward passes.
\newblock \emph{Advances in Neural Information Processing Systems}, 36:\penalty0 53038--53075, 2023.

\bibitem[Roemmele et~al.(2011)Roemmele, Bejan, and Gordon]{roemmele2011copa}
Melissa Roemmele, Cosmin~Adrian Bejan, and Andrew~S Gordon.
\newblock Choice of plausible alternatives: An evaluation of commonsense causal reasoning.
\newblock In \emph{Commonsense 2011}, 2011.

\bibitem[Su et~al.(2024)Su, Liu, Qiu, Liu, and Xu]{su2024defense}
Junda Su, Zirui Liu, Zeju Qiu, Weiyang Liu, and Zhaozhuo Xu.
\newblock In defense of structural sparse adapters for concurrent llm serving.
\newblock In \emph{Findings of the Association for Computational Linguistics: EMNLP 2024}, pages 4948--4953, 2024.

\bibitem[Wang et~al.(2018)Wang, Singh, Michael, Hill, Levy, and Bowman]{wang2018glue}
Alex Wang, Amanpreet Singh, Julian Michael, Felix Hill, Omer Levy, and Samuel~R Bowman.
\newblock Glue: A multi-task benchmark and analysis platform for natural language understanding.
\newblock In \emph{Proceedings of the 2018 EMNLP Workshop BlackboxNLP}, 2018.

\bibitem[Yu et~al.(2024)Yu, Zhou, Wang, Li, and Huang]{yu2024subzero}
Ziming Yu, Pan Zhou, Sike Wang, Jia Li, and Hua Huang.
\newblock Subzero: Random subspace zeroth-order optimization for memory-efficient llm fine-tuning.
\newblock \emph{arXiv preprint arXiv:2410.08989}, 2024.

\bibitem[Zhang et~al.(2024)Zhang, Li, Hong, Li, Zhang, Zheng, Chen, Lee, Yin, Hong, Wang, Liu, and Chen]{zhang2024zobench}
Yihua Zhang, Pingzhi Li, Junyuan Hong, Jiaxiang Li, Yimeng Zhang, Wenqing Zheng, Pin{-}Yu Chen, Jason~D. Lee, Wotao Yin, Mingyi Hong, Zhangyang Wang, Sijia Liu, and Tianlong Chen.
\newblock Revisiting zeroth-order optimization for memory-efficient {LLM} fine-tuning: {A} benchmark.
\newblock In \emph{Forty-first International Conference on Machine Learning, {ICML} 2024, Vienna, Austria, July 21-27, 2024}. OpenReview.net, 2024.
\newblock URL \url{https://openreview.net/forum?id=THPjMr2r0S}.

\bibitem[Zhou et~al.(2024)Zhou, Guo, Wang, Chang, and Wu]{zhou2024survey}
Yue Zhou, Chenlu Guo, Xu~Wang, Yi~Chang, and Yuan Wu.
\newblock A survey on data augmentation in large model era.
\newblock \emph{arXiv preprint arXiv:2401.15422}, 2024.

\end{thebibliography}

\newpage
\appendix
\section*{Appendix}
\section{Prompts for OAT-Rephrase}\label{sec:app_prompt}

We present the prompts in each phase of OAT-Rephrase for COPA dataset as an illustration.

\begin{figure*}[htbp]
    \centering
    \includegraphics[width=0.85\textwidth]{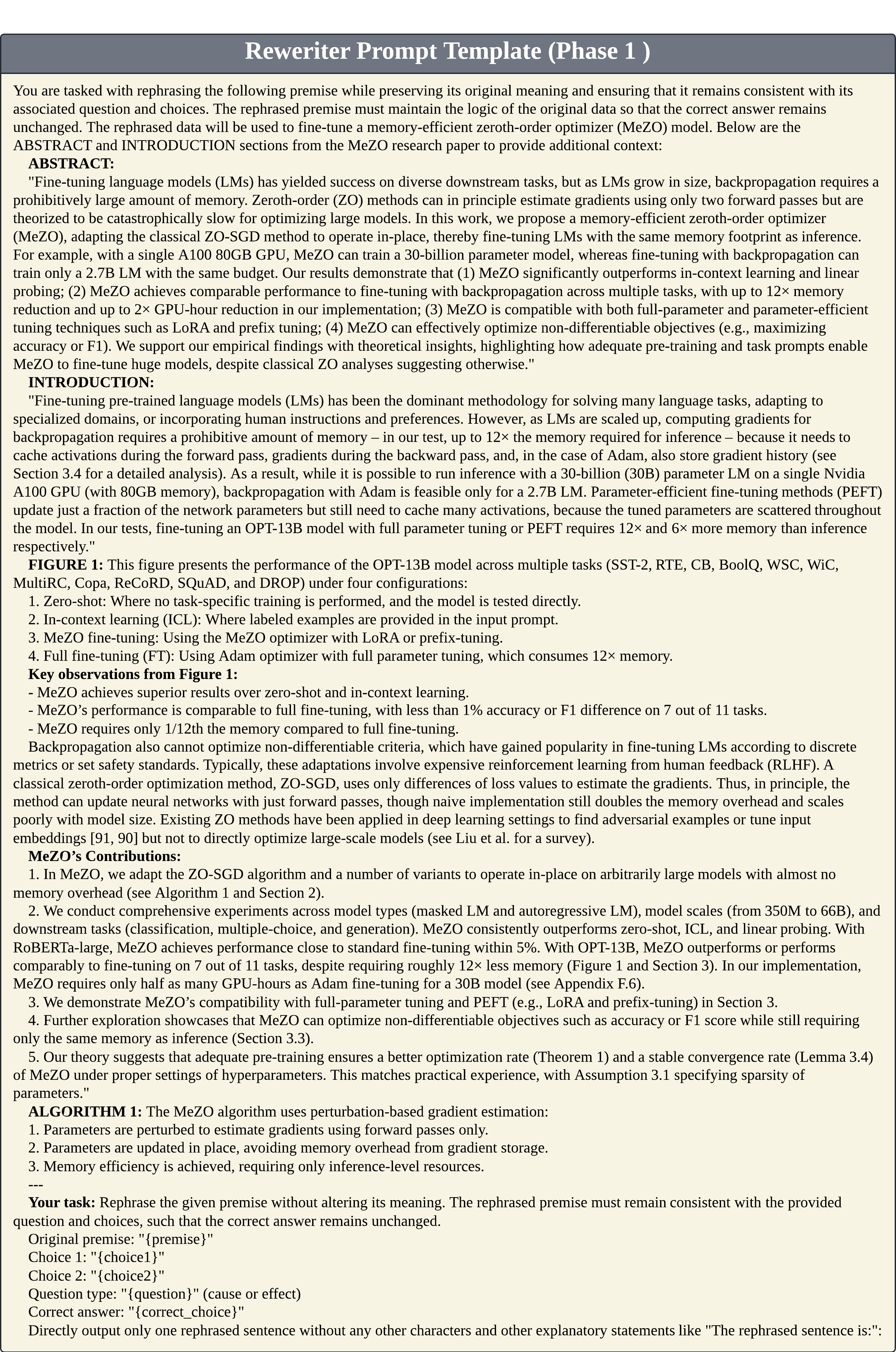}
    \caption{Rewriter Prompt Template after Phase 1.}
    \label{fig:phase1}
\end{figure*}

\begin{figure*}[htbp]
    \centering
    \includegraphics[width=0.85\textwidth]{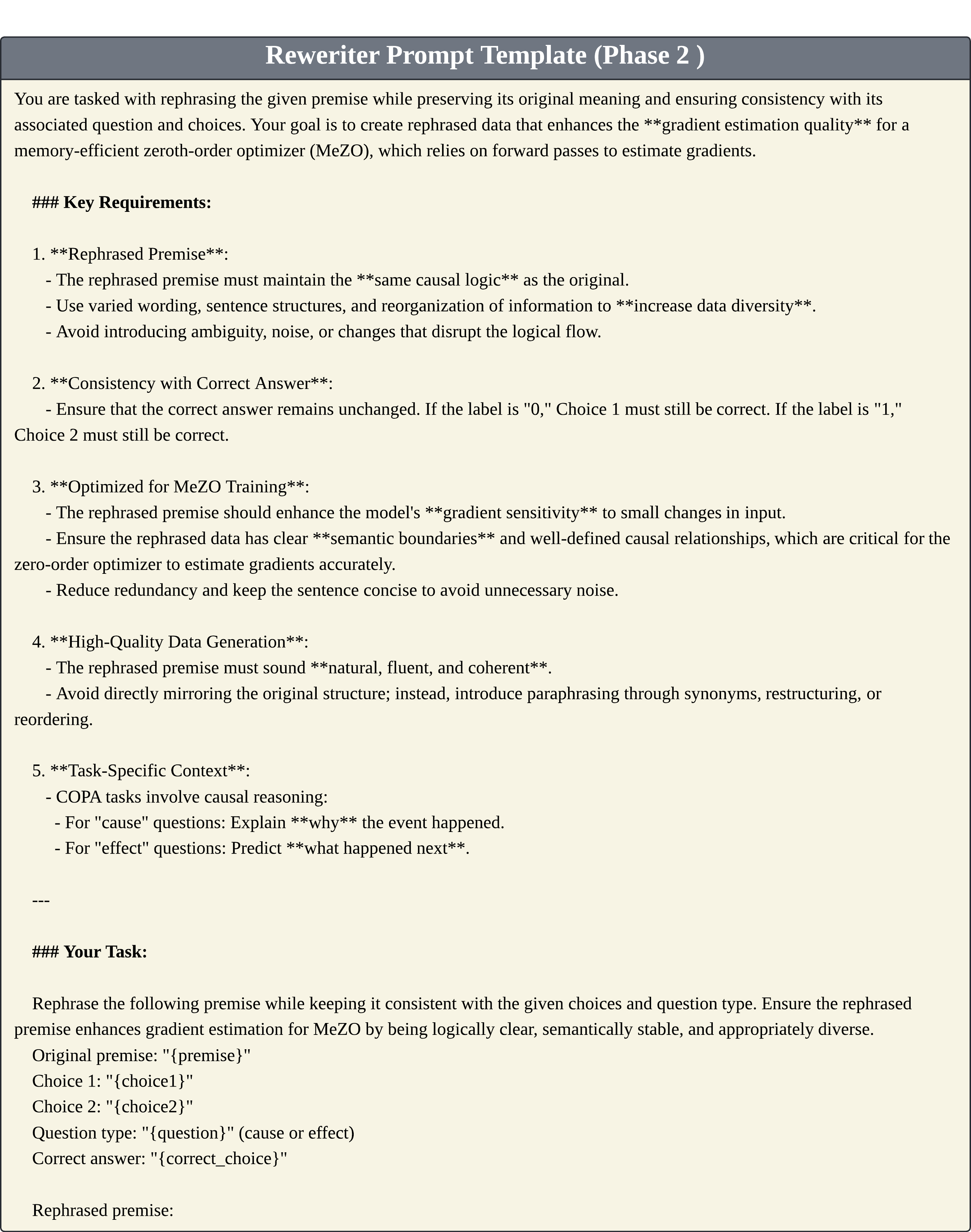}
    \caption{Rewriter Prompt Template after Phase 2.}
    \label{fig:phase2}
\end{figure*}

\begin{figure*}[htbp]
    \centering
    \includegraphics[width=0.85\textwidth]{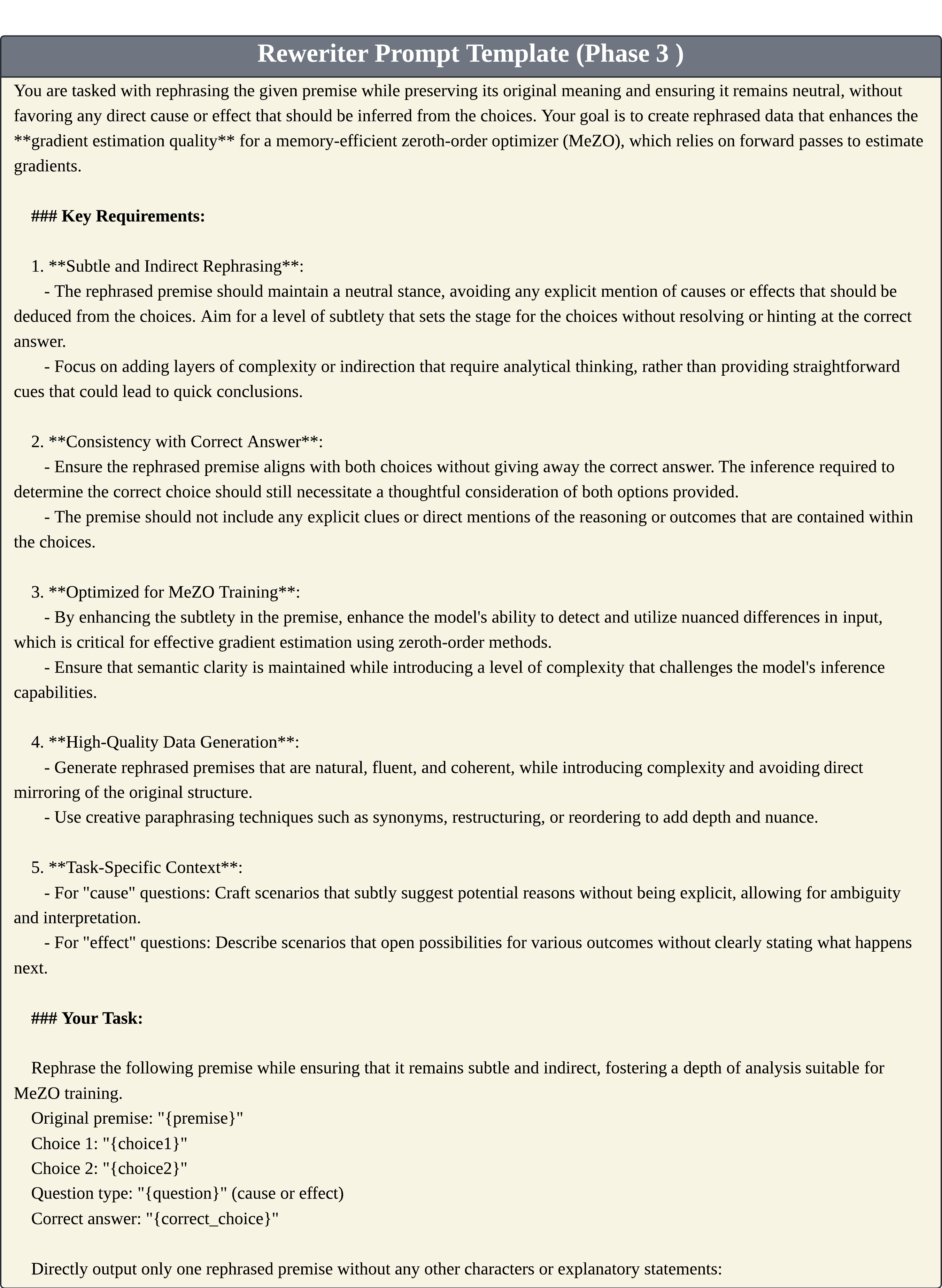}
    \caption{Rewriter Prompt Template after Phase 3.}
    \label{fig:phase3}
\end{figure*}

\begin{figure*}[htbp]
    \centering
    \includegraphics[width=0.85\textwidth]{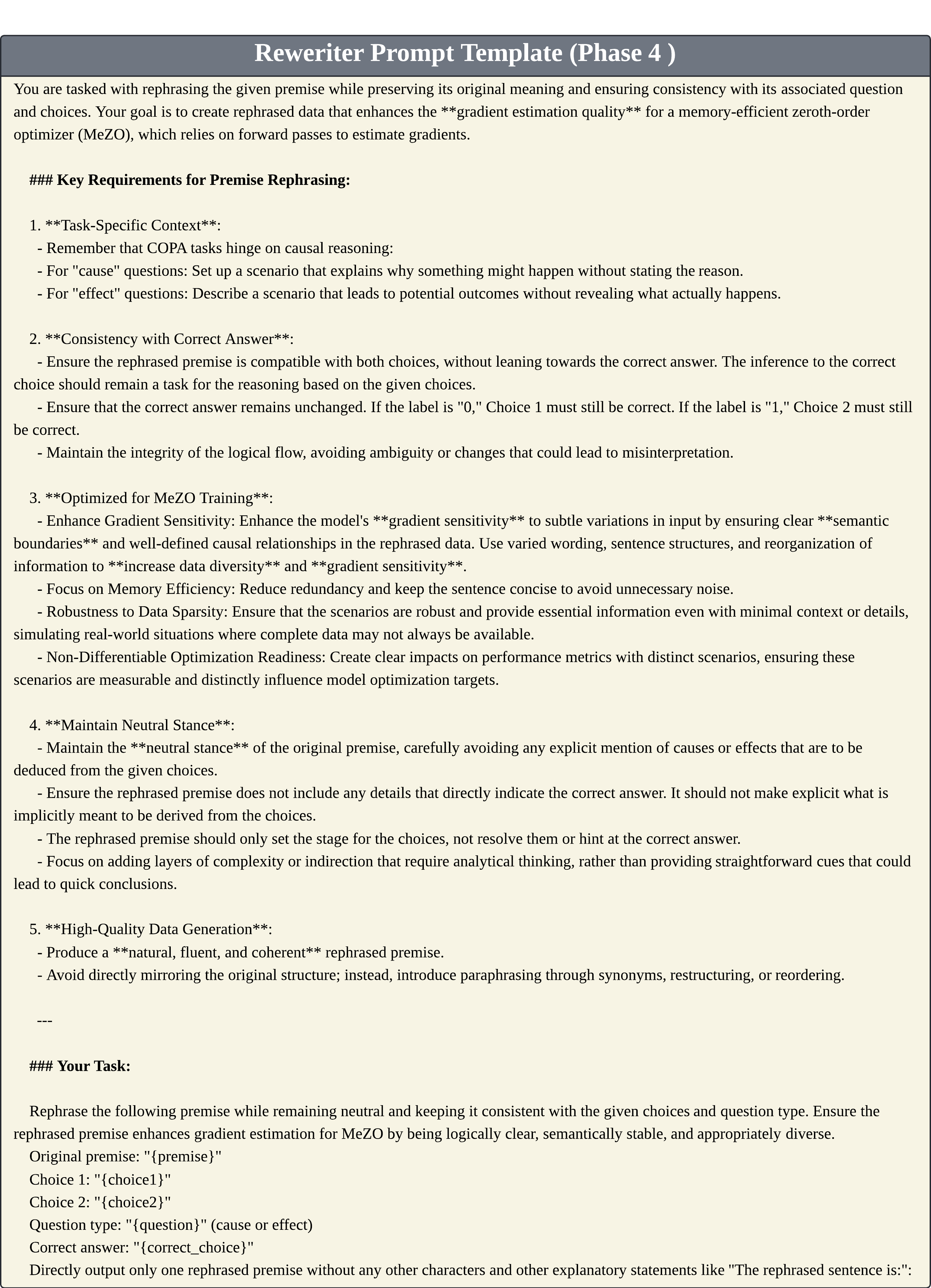}
    \caption{Rewriter Prompt Template after Phase 4.}
    \label{fig:phase4}
\end{figure*}

\begin{figure*}[htbp]
    \centering
    \includegraphics[width=0.85\textwidth]{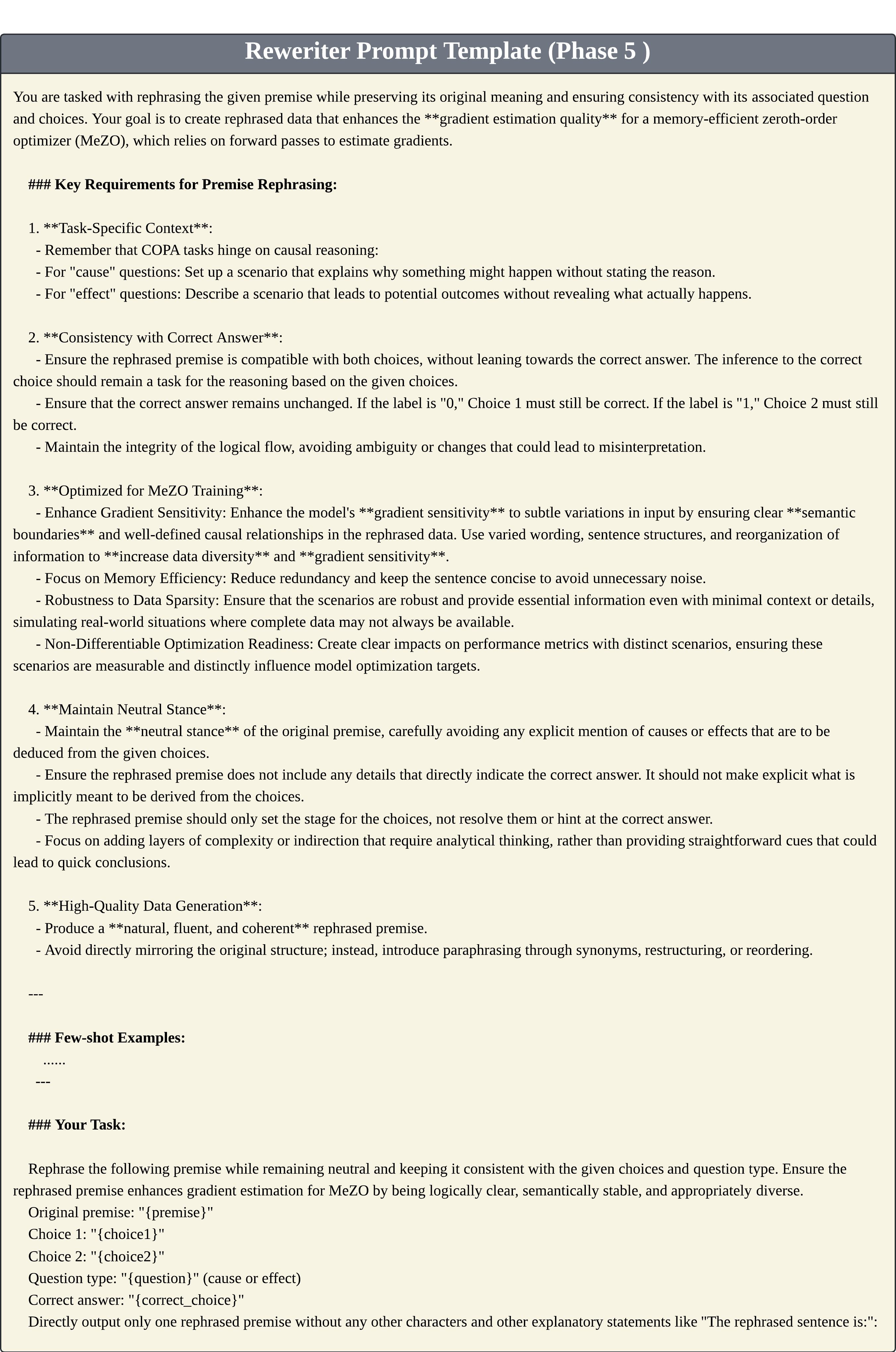}
    \caption{Rewriter Prompt Template after Phase 5.}
    \label{fig:phase5}
\end{figure*}

\begin{figure*}[htbp]
    \centering
    \includegraphics[width=0.85\textwidth]{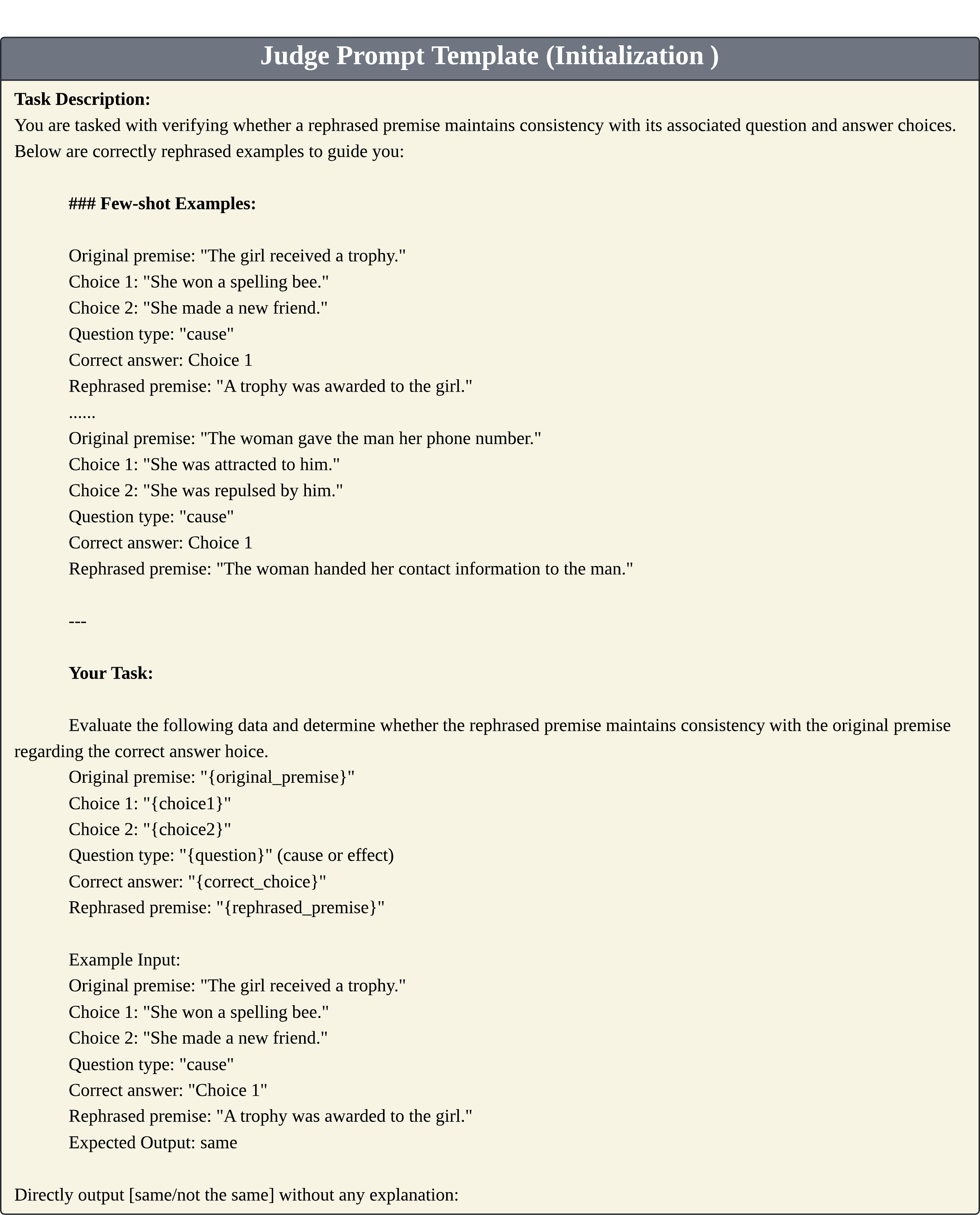}
    \caption{Judge Prompt Template Initialization.}
    \label{fig:judge1}
\end{figure*}

\begin{figure*}[htbp]
    \centering
    \includegraphics[width=0.85\textwidth]{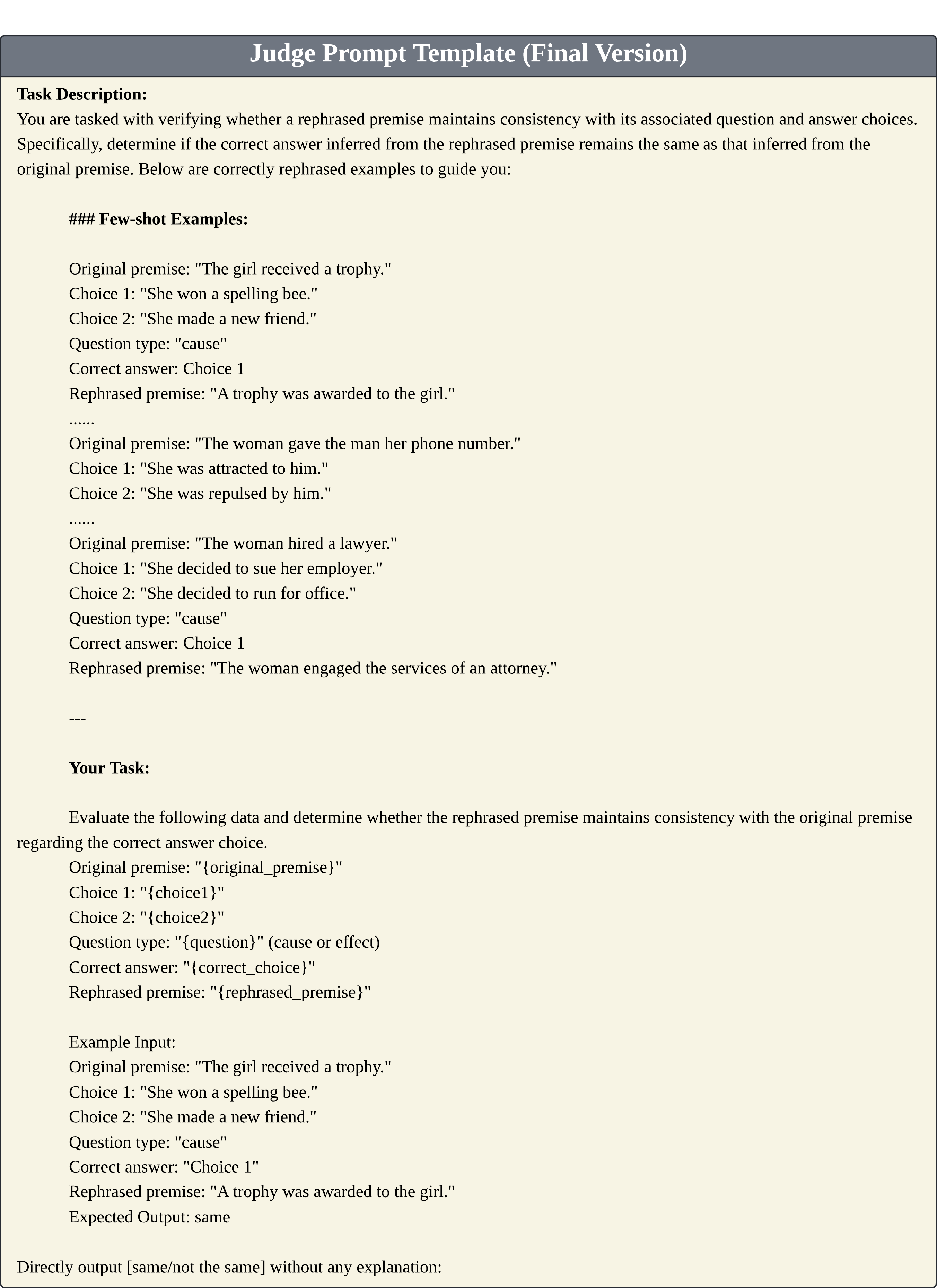}
    \caption{The Final Version of the Judge Prompt Template.}
    \label{fig:judge2}
\end{figure*}

\end{document}